\documentclass[10pt]{cai}
\usepackage{subcaption}
\usepackage{algorithm}
\usepackage{algpseudocode}

\graphicspath{{figs/}}

\begin{document}

% Editorial staff will replace the following values:
% 1. Conference Year
% 2. Issue number
% 3. Article DOI
%\volumeheader{36}{0}%{00.000}
\begin{center}

  \title{TransferD2: Automated Defect Detection Approach in Smart Manufacturing using Transfer Learning Techniques}
  \maketitle

  \thispagestyle{empty}

  % Add Authors and Affiliations in the camera ready
  % for the double blind review, please leave this section as is 
  \begin{tabular}{cc}
    Atah Nuh Mih\upstairs{\affilone}, Hung Cao\upstairs{\affilone,*}, Joshua Pickard\upstairs{\affiltwo}, Monica Wachowicz\upstairs{\affilone,\affilthree}, Rickey Dubay\upstairs{\affilfour}
   \\[0.25ex]
   {\small \upstairs{\affilone}Analytics Everywhere Lab, University of New Brunswick, Canada} \\
   {\small \upstairs{\affiltwo}Eigen Innovation Inc. } \\
   {\small \upstairs{\affilthree}RMIT University, Australia} \\
   {\small \upstairs{\affilfour}Mechanical Engineering Department, University of New Brunswick, Canada} \\
  \end{tabular}

%  Replace with corresponding author email address
  \emails{
    \upstairs{*}Corresponding Author: hcao3@unb.ca \\
    \upstairs{$\bigotimes$} This is a preprint  
    }
  \vspace*{0.2in}
\end{center}

\begin{abstract}
Quality assurance is crucial in the smart manufacturing industry as it identifies the presence of defects in finished products before they are shipped out. Modern machine learning techniques can be leveraged to provide rapid and accurate detection of these imperfections. We, therefore, propose a transfer learning approach, namely TransferD2, to correctly identify defects on a dataset of source objects and extend its application to new unseen target objects. We present a data enhancement technique to generate a large dataset from the small source dataset for building a classifier. We then integrate three different pre-trained models (Xception, ResNet101V2, and InceptionResNetV2) into the classifier network and compare their performance on source and target data. We use the classifier to detect the presence of imperfections on the unseen target data using pseudo-bounding boxes. Our results show that ResNet101V2 performs best on the source data with an accuracy of 95.72\%.  Xception performs best on the target data with an accuracy of 91.00\% and also provides a more accurate prediction of the defects on the target images. Throughout the experiment, the results also indicate that the choice of a pre-trained model is not dependent on the depth of the network. Our proposed approach can be applied in defect detection applications where insufficient data is available for training a model and can be extended to identify imperfections in new unseen data. 
\end{abstract}

% add your keywords
%\copyrightnotice

\section{Introduction}
\label{intro}
%Below is the outline of the introduction section:\\\\
%1. a paragraph describing the context of this research: We might talk about industry 4.0; smart manufacturing; automatic defection detection in the industry (why we need it)\\

The manufacturing industry is rapidly shifting to the new revolution wave (e.g., Industry 4.0 in Germany, Industrial Internet in the US) with many disruptive technologies that support effective and accurate engineering decision-making through the convergence of a vast amount of networked data and emerging technologies such as Internet of Thing, Cyber-Physical System, Edge Computing, Advanced Analytics and AI for governing and operating manufacturing operations. Many companies today are optimizing their systems by adopting more automated systems in their product manufacturing processes than by using human resources. Product quality assurance is one of the stages that are being drastically transformed into automation in the manufacturing process. With the appearance of advanced techniques in machine vision, many product imperfections can be quickly and precisely detected without the task of judging the human eye. In some cases, conventional machine vision systems can increase manufacturing efficiency and provide more accurate quality inspection with the support of several special sensors. Some examples of surface defect detection applications are Print Circuit Board (PCB) defect detection \cite{ding2019tdd}, leather defect classification \cite{Liong2020}, and steel defect detection \cite{Boikov2021}. 

\begin{figure}[!ht]
\centering
\includegraphics[width=1.0\linewidth]{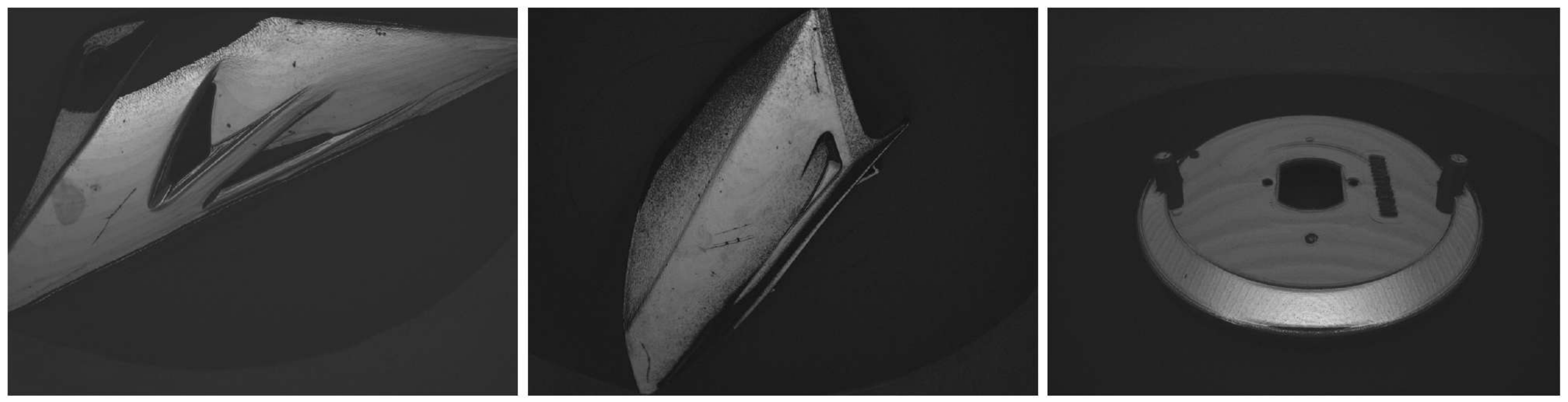}
\caption{Paint inspection images obtained with deflectometry inspection technology for three different objects.}
\label{fig_deflectometry}
\end{figure}

%2. Problem statement\\
In this work, we aim to solve complex quality inspection problems within the manufacturing industry through the application of customized deep learning models that leverage machine vision and process data, combined with advanced image processing and data standardization pipelines, to predict the existence and location of relevant quality defects. Using the deflectometry technology developed by our industry partner, a wide range of objects with specific defects can be detected in multiple applications. Figure \ref{fig_deflectometry} depicts several examples of paint inspection images applied to this technology to different objects. 

%3. What are the challenges of this work?? or What are the challenges of the problem??\\\\
Although the current inspection technology has been proven to work, it is inefficient when trying to scale an existing solution to new applications with new objects. A primary challenge when developing solutions for new applications, which may include new part geometries, new camera perspectives, additional defects, and other process changes, is ensuring Machine Learning (ML) model robustness and reliability without the need for collecting excessively large labeled datasets that deter adoption and prolong return on investments. To overcome this challenge, it is necessary to have a solution that does not require too much labeled input data but still ensures the accuracy of the trained ML model when applied to the defect identification of new objects. This requires leveraging the results of the previous solution to transfer knowledge to a new and similar application (ie. model architecture, trained models, labeled datasets). A relevant example of this challenge is paint inspection where the inspection images have well-defined features associated with specific defects (eg. scratches, contamination, paint runs, etc.) but the wide range of part geometries, colors, material, and variation in part placement present difficulties in successfully scaling a solution to new applications with new objects.

%4. (Maybe, the research objectives and research questions of this project)\\\\
The primary objective of this research project is to identify the relevance of leveraging transfer learning techniques and synthetic data approaches to minimize the necessity for large labeled datasets at the start of each new application on new objects. By using transfer learning, we aim to get a high accuracy level on the imperfection detection of new objects when we only use the learned knowledge and a limited amount of data from the initial objects. 

%5. Summarize the contributions of this paper/this work\\\\
The main contributions of this work are: 
\begin{enumerate}
    \item We propose an end-to-end approach, namely Transfer Defect Detection (TransferD2) to leverage different Transfer Learning models through an architecture design for product defect detection applications in the context of Smart Manufacturing.
    \item We devise a method to process the original small dataset and enhance the data records to produce appropriately large labeled datasets that are suitable for training robust ML models. 
    \item We also do an experiment on different sets of Transfer Learning models  representing 3 depth levels of neural network architecture (i.e., network depth level $\sim$ 100, $\sim$ 200, $\sim$ 400) and compare the performance of these models.  
\end{enumerate}

%In many manufacturing processes, different components with varying shapes, colours, and designs are usually produced. However, the types of defects remain the same across the components. Building a model for each of the components produced will be a time-consuming process and also be challenging due to the lack of data for all the components. It will therefore be optimal to build a single model that identifies the defects on all the components, irrespective of their shape, colour, and size.

%We therefore propose an approach to build a robust model that trains on limited data and is able to perform inference on components with different physical appearances. Our approach generates a large dataset from a small amount of sample data of a specific source component by splitting each image into a grid, and saving each grid tile as an independent image. By using this method, the tiles eliminate the overall component geometry and focus on the surface features to detect the defects. This approach generates a dataset large enough to train a robust model that performs well on the source component and is able to identify defects on target components that were not used during training. We proof the success of this approach by training our dataset using popular pre-built models using transfer learning. 

%6. Last paragraph is to outline the structure of this paper. 
The remainder of this paper is as follows. Section 2 describes the related work of product defect detection in smart manufacturing applications and the state-of-the-art Transfer Learning techniques. Section 3 presents our proposal for product quality assurance and imperfection detection leveraging Transfer Learning methods. Section 4 realizes our proposed approach through the implementation of a practical application on a real dataset. The evaluation and performance of our proposed approach are discussed in Section 5. Finally, we conclude our paper in Section 6.

\section{Related work}
This section discusses the state-of-the-art works related to the imperfection detection application in the manufacturing industry. We also discuss several related transfer learning approaches in the context of smart manufacturing. 
\subsection{Smart Manufacturing}
Extensive research has been done on integrating deep learning techniques in various areas of smart manufacturing, such as product quality inspection, fault diagnosis, and defect prognosis \cite{Wang2018}. In our paper, we explore product quality inspection and focus on surface defect detection. This research area leverages image processing techniques and computer vision to classify and localize defects on the surface of materials.  

One of the major challenges faced by deep learning in surface defect detection has been the lack of large amounts of training data necessary for training deep learning models \cite{Zhu2021}. Several authors have explored various approaches to deal with this challenge. For example, Tabernik et al. \cite{Tabernik2020} proposed a segmentation-based architecture for the detection and segmentation of surface defects. In their method, each pixel of the surface defect was considered a training sample, thereby increasing the amount of training data available. In another work \cite{Liong2020}, the authors had 27 sample images of leather surface defects. They proposed a method to split each image into 24 smaller ones but manually selected images such that each image only contained one defect. Another research \cite{Boikov2021} proposed generating synthetic data for steel defect detection using rendering software to re-create the images. The images were shifted to be viewed from different angles, and shader parameters were altered to vary the defect shapes and locations. Jain et al. \cite{Jain2022} also proposed using a Generative Adversarial Network to generate synthetic data for surface defect detection. They used a generator network to produce realistic synthetic images, which were then used to train their classifier.    

The nature of modern manufacturing systems provides flexible configurations for producing a wide variety of products. Using traditional machine learning methods will pose a challenge as the feature representations will need to be redesigned from scratch. To overcome the challenge of domain shift, several transfer learning techniques have been proposed.  

\subsection{Transfer Learning}
Transfer learning has emerged as a popular deep learning approach in which knowledge from a \textit{source domain} is applied to a \textit{target domain} to improve training performance. This approach has become popular in applications in which large training data is not available to build a model from scratch. By using transfer learning, the parameters from the pre-trained model can be adjusted to suit the new domain which provides faster convergence than would have been achieved from random weight initialization when training from scratch.  

A popular source domain for transfer learning has been the ImageNet dataset. The open-source dataset consists of over 14 million images, spanning 1000 classes of objects, and has been used as a reference by researchers for their deep learning models. Many deep learning models developed for this dataset have become base models for transfer learning in various applications such as in garbage classification \cite{Rismiyati2020}, fault detection in rail components 
\cite{Yilmazer2022}, fault diagnosis \cite{Wen2019}, diagnosing and anticipating leukemia \cite{Abir2022}.  

Specific to smart manufacturing, several frameworks have also been proposed that build upon transfer learning. Zhu et al. \cite{Zhu2020} used transfer learning to build a model that could detect bridge defects in a dataset of images not used during training. They used InceptionV3 as a feature extractor and built a classifier network on top of the base model. Gong et al. \cite{Gong2020} proposed a deep transfer learning method to identify defects in aeronautics composite materials. They trained the labeled images of their source and target data together by feeding image pairs consisting of both images from both datasets as image pairs, thereby providing large enough input data for training their model.  

The choice of the base model when building a classifier for a new dataset plays an important role \cite{Zhu2021}. Abu et al. \cite{Abu2021} did a comparative study on the performance of ResNet, VGG, MobileNet, and DenseNet on the SEVERSTAL and NEU datasets. They used transfer learning to train the models on these datasets and found MobileNet to be the best-performing model. In another study with a different dataset \cite{Zhao2021}, the researchers found Xception to be the most performant model on the Environmental Microorganism Image Dataset (EMDS-6) dataset. Another study \cite{Zhu2021} chose ResNet-18 as the base model of their study due to its relatively simple implementation and high performance. This proves that there is no clear metric in selecting the base model for implementation. 

% \subsection{Few Shots Learning}

% \subsection{Self-supervise Learning}
% Self-supervised learning is a machine learning in which machines learn by observation and figure out the underlying structure of the images, speech, or text. This approach provides a solution to the problem of unlabelled data in machine learning by leveraging the data's co-occurrence relationships. 

\section{Proposed Method}
\subsection{Overview approach} \label{overview_approach}
The detection of surface defects can be considered a binary image classification problem \cite{Tabernik2020}, in which the image analyzed is either defective or non-defective. Emphasis should therefore be laid on identifying features that enable the model to correctly classify an image as defective or not. This implies irrelevant features such as the object’s shape, size, and color should be given less importance in the classification of the image. As such, a model trained to correctly identify surface defects on a specific object will therefore have a significant performance when identifying defects on a new object, irrespective of shape, size, or color.    

\begin{figure}[!ht]
\centering
\includegraphics[width=\linewidth]{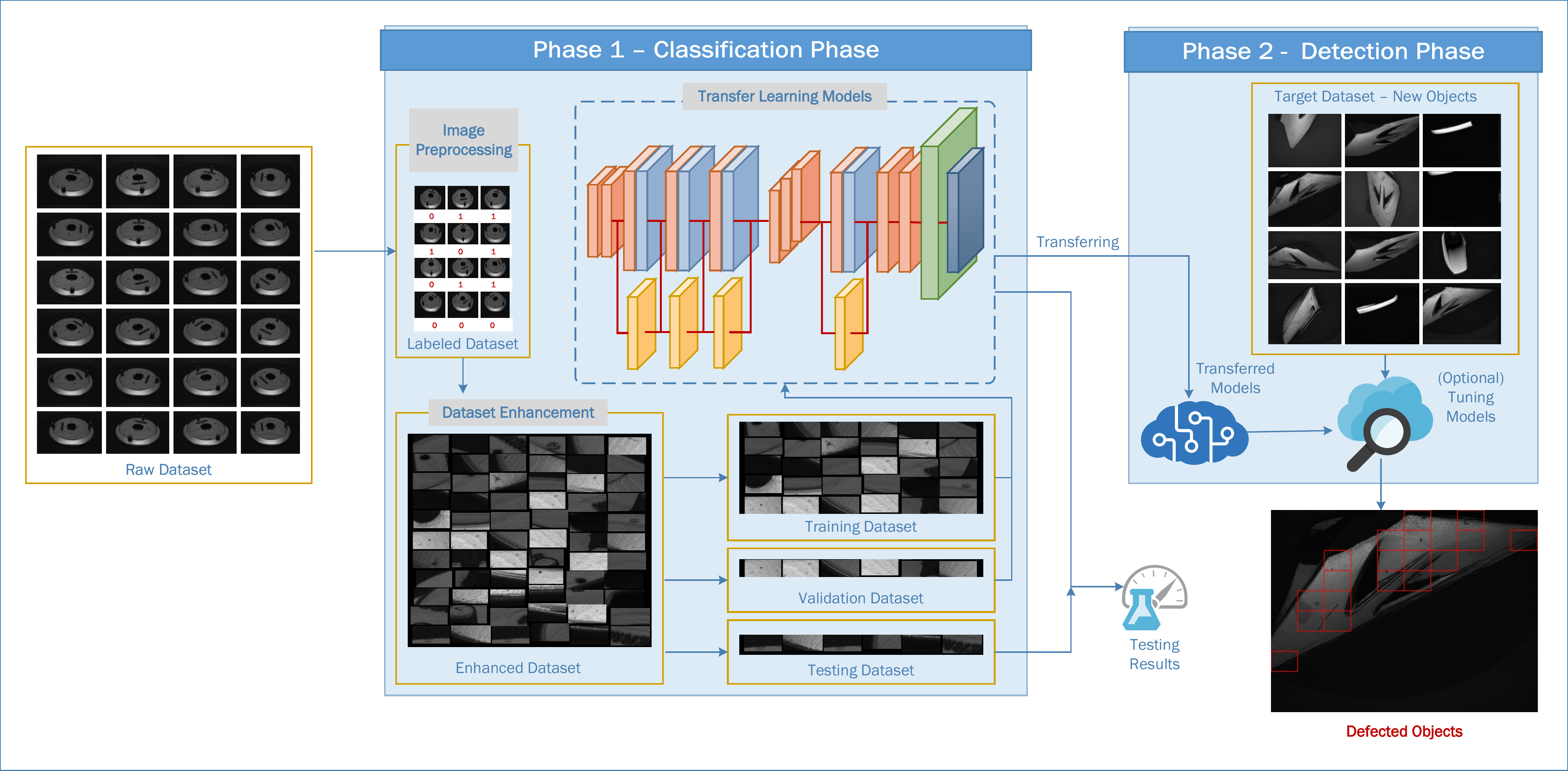}
\caption{The proposed methodology for product imperfection detection using Transfer Learning.}
\label{fig_overall_approach}
\end{figure}

We propose an approach, as shown in Figure \ref{fig_overall_approach}, that is divided into two phases: a classification phase and a detection phase. In the classification phase, we build a binary classifier that identifies if an image is defective or not.  The detection phase then uses the classifier to run inference on new data and identify the defective segments of the image. We explain both phases in detail in the following subsections. 

\subsection{Classification Phase} 

The execution of an accurate classifier is the most crucial step in our proposed approach. The classification phase is further broken down into Data Pre-processing, Dataset Enhancement, and Training and Building of Learning Models stages. 

\subsubsection{Data Pre-processing}
Data pre-processing provides the basis for building the classifier’s dataset. In this stage, each image in the source dataset is split into smaller tiles, and the tiles are labeled as defective or non-defective. 

The image is split into a grid of \textit{m} columns and \textit{n} rows. Tiles are labeled as either defective or non-defective by verifying if they contain a defect. If the coordinates of the defect on the original image overlap or lie within the coordinates of the tile, that tile is marked as defective. Otherwise, the tile is marked as non-defective. The \textit{Label}, \textit{Source Image ID}, and \textit{x} and \textit{y} values of the tile are used to save the tile as an independent image in the new dataset (where \textit{Label = 0} for non-defective and \textit{Label = 1} for defective), thereby providing a unique file name for each tile. Algorithm \ref{alg:cap1} presents the procedure to pre-process all images in the Source Dataset, \textit{S} and generate a new Enhanced Dataset, \textit{E} by cropping the original images and splitting them into labelled tiles. Specifically, each original source image in the pre-processing stage is split into a grid of \textit{m} columns and \textit{n} rows, and each tile is saved as a new image. This implies, each original source image produces \textit{m*n} tiles that are saved as independent images. 

Consider a source dataset of \textit{N} images. After pre-processing all the images in the source dataset, we obtain a new dataset of size \textit{E = N*m*n} images and their corresponding labels (0 for non-defective and 1 for defective). The detailed implementation for this procedure is explained in Section \ref{subsec:data_preprocessing}.  

The classifier’s performance depends on its identification of crucial features of the image, which include the defect itself and its surroundings. The classifier is trained to learn features that identify if a tile is defective or not rather than learning the general features of the object itself. Hence, splitting the image into tiles provides a basis for the generalization of the classifier on a dataset with unseen and unlabelled images.   

\begin{minipage}{0.46\textwidth}
\begin{algorithm}[H]
\caption{Dataset Pre-processing}\label{alg:cap1}
\textbf{Input:} \text{Source Dataset, S} \hspace*{\fill} \\
\textbf{Output:} \text{Enhanced Dataset, E}\hspace*{\fill} 
\begin{algorithmic}[1]
\State{Let E be an empty dataset}
\For{$img \in S$}
    \State $ img \gets crop(img) $
    \For{$i \gets 0, m$} \Comment{m columns}
        \For{$j \gets 0, n$} \Comment{n rows}
            \State $tile \gets img[i:j]$
            \If{tile has defect}
                \State $label \gets 1$
                \State $E \gets tile, label$
            \Else
                \State $label \gets 0$
                \State $E \gets tile, label$
            \EndIf
        \EndFor
    \EndFor
\EndFor

\end{algorithmic}
\end{algorithm}
\end{minipage}
   \hfill
\begin{minipage}{0.46\textwidth}
\begin{algorithm}[H]
\caption{Dataset Enhancement}\label{alg:cap2}
\textbf{Input:} \text{Enhanced Dataset, E} \hspace*{\fill} \\
\textbf{Output:} \text{Balanced Dataset, D}\hspace*{\fill}
\begin{algorithmic}[1]
\State{Let D be an empty dataset}
\State{Let class 0 be an empty set of non-defective tiles}
\State{Let class 1 be an empty set of defective tiles}
\For{$index, tile, label \in E$}
    \If{index is even}
        \State{select tile with label = 1}
        \State{randomly rotate or flip tile}
        \State{assign tile to class 1}
    \Else
        \State{select tile with label = 0}
        \State{randomly rotate or flip tile}
        \State{assign tile to class 0}
    \EndIf
\EndFor
\State{$D \gets Class0 + Class 1$}
\end{algorithmic}
\end{algorithm}
\end{minipage}

\subsubsection{Dataset Enhancement} 
\label{subsec:dataset_enhancement}
Defects occupy a small area of the image, and defective images will, therefore, provide a significantly smaller percentage of the images in the new dataset E. Such data imbalance will result in poor model performance \cite{Rismiyati2020} as training will be biased towards the class with a more significant number of samples \cite{Saini2019}. Several methods have been proposed to solve the data imbalance problem, such as an even-odd mechanism to balance the defective and non-defective samples \cite{Tabernik2020}.   

In the dataset enhancement stage, we also use an even-odd mechanism to select images and oversample the class of defective images randomly. Algorithm \ref{alg:cap2} illustrates our approach to obtain a balanced dataset D from an enhanced dataset E in previous stage. At even iterations, random defective samples are selected and randomly rotated or flipped to avoid repetitive images that could lead to overfitting. The same approach is used to select random non-defective samples at odd iterations. This process results in the oversampling of defective images and the undersampling of non-defective images. Extensive training data is required to tune the parameters of the transfer learning model; therefore, we iterate the data balancing process over the size of the imbalanced dataset to produce a balanced dataset of the same size. The resulting dataset is evenly balanced with an equal number of defective and non-defective samples. The images are assigned to their respective classes, that is, 0 for non-defective samples and 1 for defective samples, and the dataset is then split into the training and validation datasets.

% \begin{minipage}{0.46\textwidth}
% \input{contents/algorithm/Alg1.tex}
% \end{minipage}
% \hfill
% \begin{minipage}{0.46\textwidth}
% \begin{algorithm}[H]
%     \centering
%     \caption{Example Algorithm}\label{algorithm1}
%     \begin{algorithmic}[1]
%         \State \text{dataProcessing (dataset)} 
%         \State $\textit{unitToDrop} \gets \text{25\%}$
%         \State \text{text}
%         \State \text{text}
%         \Repeat
%         \State \text{/*text*/}
%         \For{$i\gets 1, rows$}
%         \State $\text{text}$ 
%         \State $\text{text}$ 
%         \State $\text{text}$ 
%         \State \text{text}
%         \State $\text{text}$
%         \EndFor
%         \Until {text}
%         \State \text{text}
%         \Repeat
%         \State \text{text}
%         \For{$i\gets 1, rows$}
%         \State $\text{text}$ 
%         \State $\text{v}$ 
%         \EndFor
%         \Until {text}
%         \State \textbf{Return}  \text{text, text}
%     \end{algorithmic}
% \end{algorithm}
% \end{minipage}

\subsubsection{Training and Building of Learning Models} 

Transfer learning has been proven to be beneficial over training a model from scratch \cite{Hussain2019} by making use of predefined weights from the pretrained model. These predefined weights are then adjusted to learn features on a new dataset \cite{Hussain2019}, thereby saving computational time required to train a model from scratch with random weight initialization.  

Our proposed approach uses transfer learning to build the classifier network on top of a model whose weights are pre-trained on the ImageNet dataset. The classifier network consists of an average pooling layer, a dropout layer, and the output layer is a dense layer with two neurons (for binary classification) and a sigmoid activation function. To obtain better performance, the entire network is trained end-to-end, and binary cross entropy is used as the loss function.

\subsection{Detection Phase} \label{proposed_defect_detection}

The saved transfer learning models are then loaded for inference on the target dataset containing unlabelled images of different objects. The classifier is trained to classify tiles from input images and not the entire image itself.  As the classifier is built using a sigmoid function, it returns prediction values within the range of 0 to 1. The closer the prediction is to 0, the more likely it is to be a non-defective image, and the closer the image is to 1, the more likely it is to be a defective image. This implies that prediction values greater than 0.5 are considered defective and vice versa. To prevent the mislabelling of defective tiles, the prediction threshold should be kept suitably higher than 0.5. 

A sliding window is used to split the image of the new object from the target dataset into a grid of tiles, and each tile is input into the classifier for prediction. The result of the prediction is then used to determine if the tile is defective or not, as described above. Defective tiles are then flagged and bounded by a rectangle on the input image, thereby providing pseudo-object detection.    

\section{Implementation}
\subsection{Deflectometry and data acquisition}
% @JOSH: PLEASE PROVIDE A LITTLE BIT MORE INFORMATION ABOUT THE SENSORS, DEFLECTOMERY TECHNOLOGY AND HOW WE OBTAIN THE DATASET.
Deflectometry inspection systems use synchronized patterned lighting and image capture, combined with image processing and deep learning, to inspect high gloss parts for surface and subsurface defects. Here, 16 black and white patterned images are shown on a digital display, a Flir BFS-PGE-27S5M-C monochrome camera captures the 16 corresponding images of the pattern reflections on the part surface, and deflectometry image processing generates a single processed image that clearly shows the defects. This inspection is repeated multiple times for each unique part and the resulting processed images are annotated to identify all scratch and dirt/contamination defects visible in the image.

The resulting dataset contains 227 black and white images of a source object and defect annotations that define the type of defect and their bounding boxes. As mentioned in Section \ref{overview_approach}, we focus on a binary classification problem and therefore ignore the different classes of defects.

\subsection{Data Pre-processing}
\label{subsec:data_preprocessing}
\begin{figure}[!ht]
\centering
\includegraphics[width=\linewidth]{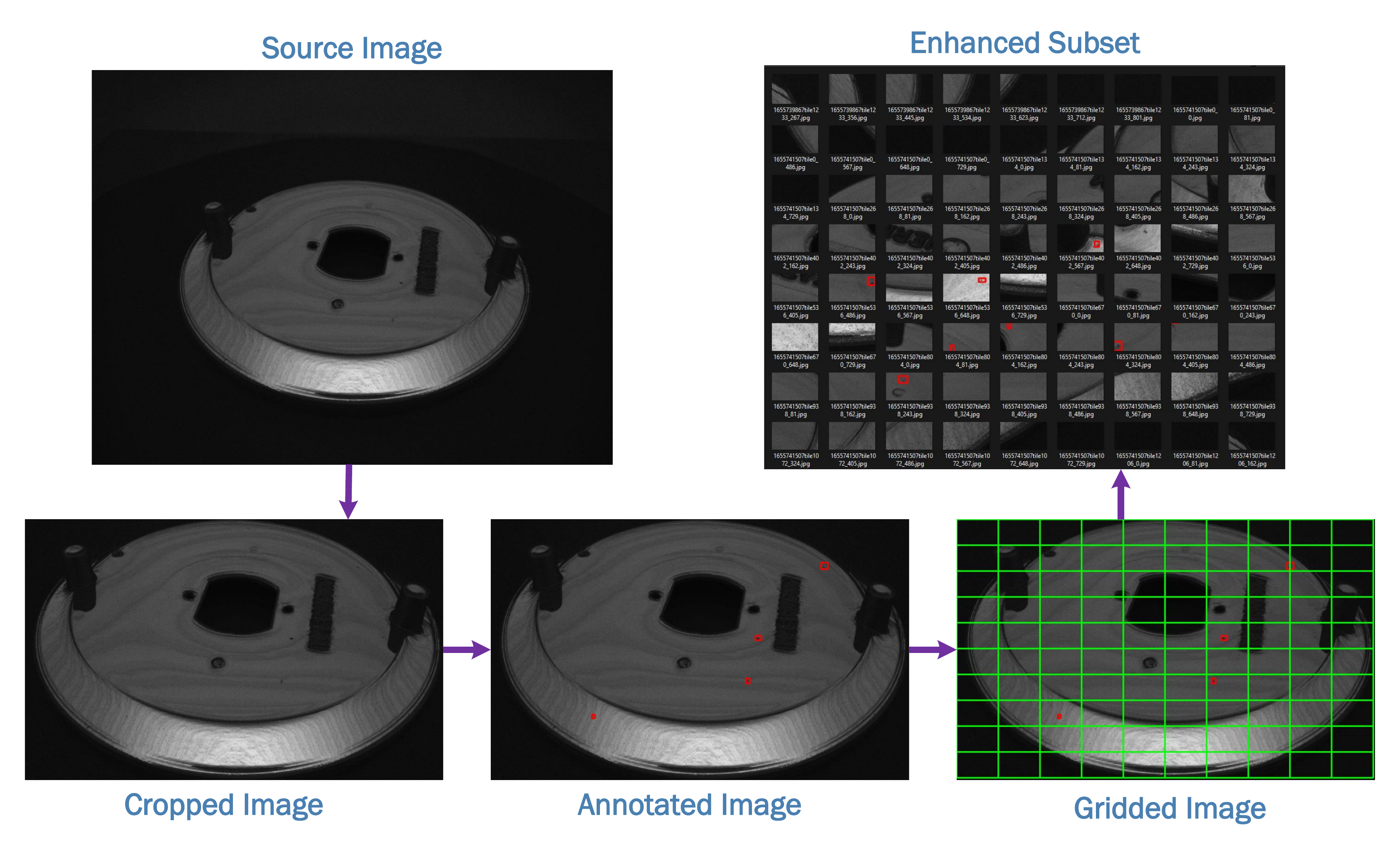}
\caption{The proposed method for dataset enhancement}
\label{fig_preprocessing}
\end{figure}

The source dataset contains images and the corresponding defect annotations. As seen from Figure \ref{fig_preprocessing}, the object is bounded within a black background, which is irrelevant for the classifier. Canny edge detection provides a better method of background removal over fixed cropping due to the varying shape of the objects in the image. With canny edge detection, an image box is dynamically identified based on the object’s edges, and the resulting box is used to crop the image. In contrast, cropping with a fixed box fails to take into consideration the dimensions of the object and crops out parts of the object that fall outside the prescribed cropping range.  

The image obtained from cropping has different dimensions from the original image, and therefore, the annotations must be mapped to the new image. The location of pixels on an image is referenced from the image’s origin coordinates. Cropping does not modify the distance of the pixels from the origin, but rather re-assigns a new point of origin. Therefore, the coordinates of the annotated pixels can be mapped to the new image by calculating new coordinates based on the cropped image’s new point of origin. The new image’s defects are now mapped from the re-assigned annotation values.  

\subsection{Dataset Enhancement}
For each image in the dataset, the black background is cropped out to prevent having a dataset of redundant black images. The annotations are mapped onto the cropped image as shown in Figure \ref{fig_preprocessing}. (Note that the boxes and grid are only for demonstration purpose). The cropped image is divided into a 10x10 grid whose tiles become the images of the new dataset, thereby generating 100 labeled images for the new dataset. Each tile is labeled as either 0 for non-defective or 1 for defective and assigned a unique file name that includes its label for easy identification.

Applying the process above on all the images in the source dataset generates a dataset of 22,700 images for both the defective and non-defective classes. However, the dataset was heavily imbalanced, with 1609 images in the defective class and 21,091 images in the non-defective class. The even-odd approach to data balancing mentioned in the previous Section \ref{subsec:dataset_enhancement} generated a balanced dataset with 11,350 images in each of the classes. To train and evaluate the model, the dataset is split into the train, validation, and test dataset in the ratio 8:1:1. 

\subsection{Transfer Learning Model Design}
\begin{figure}[!h]
\begin{minipage}[c]{.45\linewidth}
  \centering
    \includegraphics[width=\linewidth]{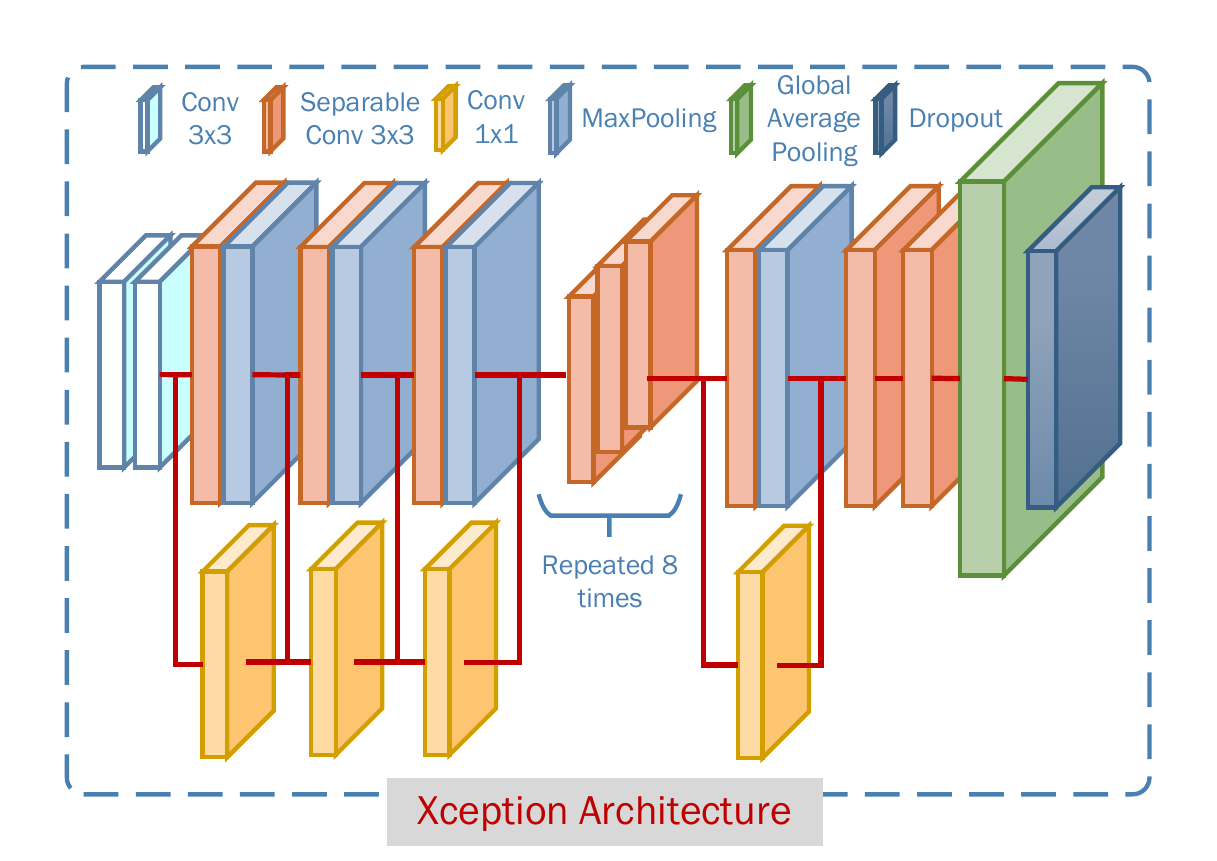}
      \subcaption{The general Xception architecture}\label{fig:4a}

  \vspace{1mm}

  \includegraphics[width=\linewidth]{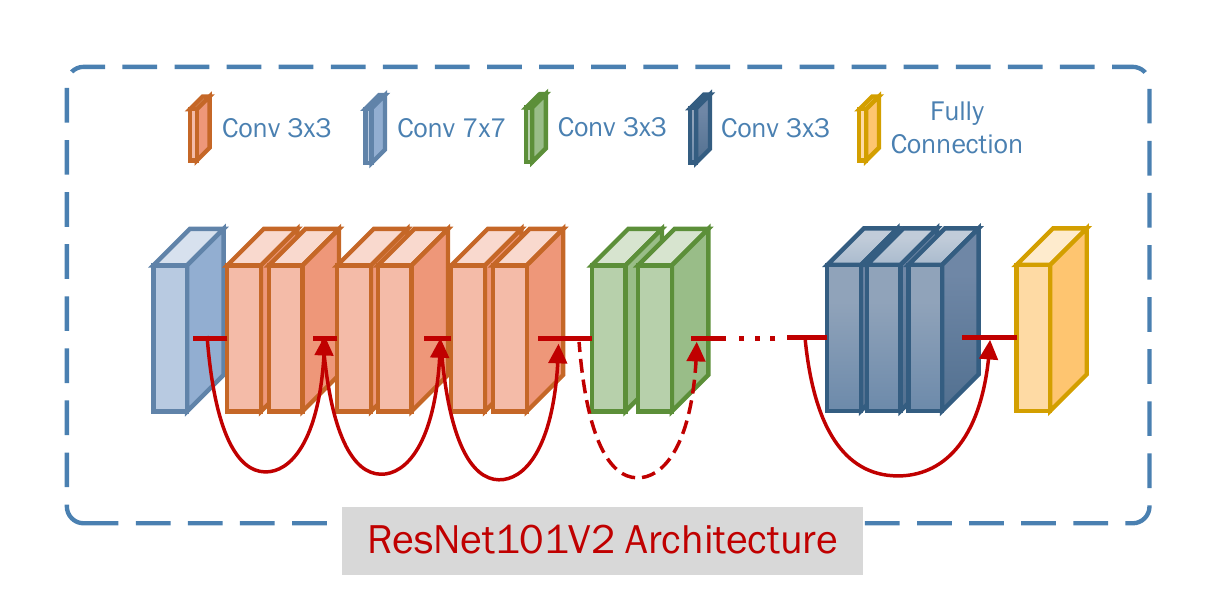}
    \subcaption{The general ResNet101V2 architecture}\label{fig:4b}
  
  \vspace{1mm}

  \includegraphics[width=\linewidth]{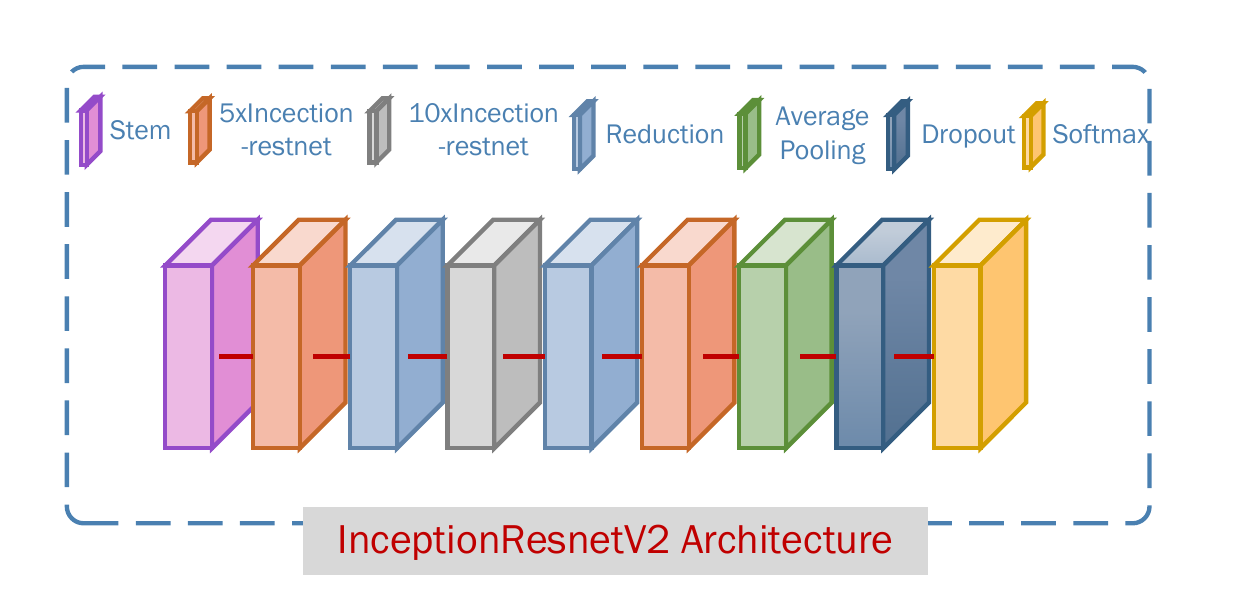}
    \subcaption{The general InceptionResNetV2 architecture}\label{fig:4c}
\end{minipage}\hfill
\begin{minipage}[c]{.5\linewidth}
  \centering
    \includegraphics[width=\linewidth]{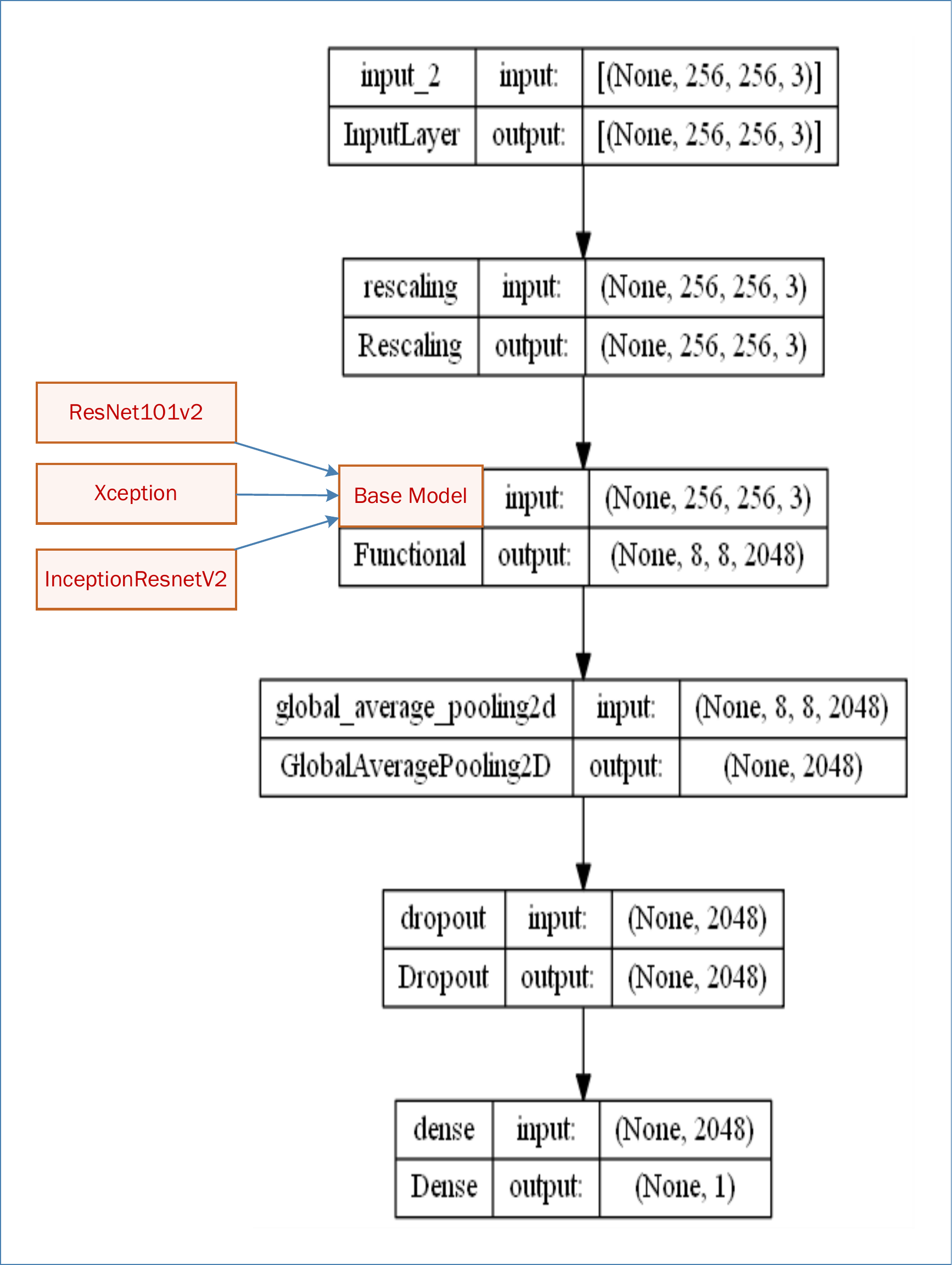}
      \subcaption{The Transfer Learning architecture}\label{fig:4d}
\end{minipage}
\caption{The proposed architecture design for training 3 different Transfer Learning models}\label{fig:4}
\end{figure}

The success of transfer learning approaches lies in the choice of the base model selected \cite{Zhu2021}. The depth of the network plays an important role in the accuracy of the CNN, as deeper networks extract more complex features about the dataset. In our implementation, the selection of the base model depends on the depth of the network. We choose a pre-trained model from each of the following three classes of models: network depth level $\sim$ 100; $\sim$ 200; and $\sim$ 400 to evaluate our approach. Figure \ref{fig:4} illustrates the architecture of our network. 

% \begin{figure}[H]
% \begin{minipage}[c]{.45\linewidth}
%   \centering
%     \includegraphics[width=\linewidth]{figs/Implementation - all - diagram.pdf}
%       \subcaption{Another subfigure}\label{fig:4a}
% \end{minipage}\hfill
% \begin{minipage}[c]{.45\linewidth}
%   \centering
%     \includegraphics[width=\linewidth]{figs/Implementation - Exception.pdf}
%       \subcaption{A subfigure}\label{fig:1a}

%   \vspace{1mm}

%   \includegraphics[width=\linewidth]{figs/Implementation - Restnet101v2.pdf}
%     \subcaption{Another subfigure}\label{fig:1b}
  
%   \vspace{1mm}

%   \includegraphics[width=\linewidth]{figs/Implementation - InceptionResnet.pdf}
%     \subcaption{Another subfigure}\label{fig:1b}
% \end{minipage}

% \caption{A figure}\label{fig:1}
% \end{figure}

In our first class of the base model, we choose Xception 
\cite{Chollet2017} with 81 convolution layers. Xception is inspired by the Inception model but is more lightweight and offers better performance (79.0\% Top-1 Accuracy on the ImageNet dataset) \cite{Chollet2017}. In our second class of the base model, we choose ResNet101V2 \cite{He2016identity}, with a network depth of 205 layers. ResNet101V2 is part of the residual family of networks that are easy to train and whose accuracy increases with an increase in depth. The model achieved a Top-1 Accuracy of 77.2\% on the ImageNet. In our final class of the base model, we choose InceptionResNetV2 \cite{Szegedy2016}, with a network depth of 449 layers. InceptionResNetV2 combines the Inception architecture and residual connections, which significantly improves the training of the Inception networks. The model achieved 80.3\% Top-1 Accuracy on the ImageNet dataset.  

We integrate the pre-trained models in a new network consisting of an Input layer, a Rescaling layer, a Global Average Pooling layer, a Dropout layer, and an Output layer. The shape of the image (fixed for all images) is specified as the parameters for the Input layer. The range of the input image’s pixel values is converted from [0,255] to [-1,1] in the Rescaling layer. This normalization ensures that large pixel values are not given more importance over smaller pixel values during weight updates. The output of the Rescaling layer is connected to the input layer of the base model. The base model then serves as a feature extractor, and its pre-trained weights provide the basis for which information can be learned about the new dataset.  

During feature extraction, all features are mapped to their corresponding feature space in a convolutional process. However, the convolution does not differentiate between relevant and redundant features, resulting in a large feature space. The Global Average Pooling layer reduces the redundant features by taking the average of each feature map and passing it to a Softmax function. The resulting feature maps, therefore, represent categories of confidence maps and are more robust to spatial translations of the input. 

During training, the model is prone to overfitting on the training dataset as some settings of weights can perfectly predict the output but fail to predict correctly on the validation data. By introducing the Dropout layer, some neurons are randomly omitted from the network and cannot be depended upon for weight update. The prevention of this strong co-dependency between neurons thus prevents overfitting. 

The output of the network is passed through a Sigmoid activation function in the Output Layer. The Sigmoid function returns the prediction as a probability value between 0 and 1. Output values closer to zero represent predictions of the non-defective class, while predictions closer to 1 represent the defective class.  

An identical network is repeated for all three base models, maintaining the overall architecture but only altering the pre-trained model. The model is then trained and fine-tuned for 50 epochs and saved for inference in the Detection Phase. 

\subsection{Defect Detection} \label{implemented_defect_detection}
During inference, a random image is loaded from the target dataset. The image is split into an equal number of tiles as specified during the pre-processing step. A sliding window selects individual tiles and passes them one by one into the model for prediction. A threshold value of 0.7 is assigned to determine if the prediction should be classified as defective or not. As greater importance is given to correctly classifying defective tiles, we set the threshold to be comfortably higher than 0.5 but not too high, as it could miss out on the classification of some images. 

A bounding box is drawn over the defective tiles on the overall image providing a pseudo-object detection. We discuss the results of the defect detection in Section \ref{results}.

\section{Results and Discussions} \label{results}
\subsection{Evaluation Metrics}
The choice of several metrics provides a more robust, accurate comparison as one classifier could perform well on a single metric and poorly on a different metric \cite{Seliya2009}. We, therefore, evaluate the performance of our classifiers using binary accuracy, F1 Score, and AUC. 

The accuracy gives the ratio of correct classifications of the model. Accuracy is obtained by comparing the predicted label against the ground truth label. A model with high accuracy correctly predicts the label of the image majority of the time. In evaluating the performance of a classifier, we also want to know the fraction of relevant instances out of the total number of predicted instances (precision) and the fraction of relevant instances that were correctly predicted out of the total relevant instances (recall). The precision gives the fraction of true positives against the number of predicted positives (i.e., the sum of true positives and false positives). Recall gives the fraction of true positives against the number of actual predicted positive instances (i.e., the sum of true positives and false negatives).   

A simple measure of the precision and recall of different classifiers could be misleading as a classifier may have better precision and worse recall as compared to another classifier. We, therefore, use the F1 Score of the models to provide a more accurate comparison of the classifiers. The F1 Score measures the harmonic mean of precision and recall. It provides a reliable metric for evaluating classifiers on a balanced dataset \cite{Chicco2020}.  

The AUC (Area Under the ROC Curve) provides an alternative measure of performance in the absence of a confusion matrix \cite{Chicco2020}. It evaluates all possible classification thresholds of a model, thereby giving an accurate quality measure of the classifier. 

\subsection{Evaluation on the Source Dataset}
\begin{figure}[h!]
     \centering
     \begin{subfigure}[b]{0.4\textwidth}
         \centering
         \includegraphics[width=\textwidth]{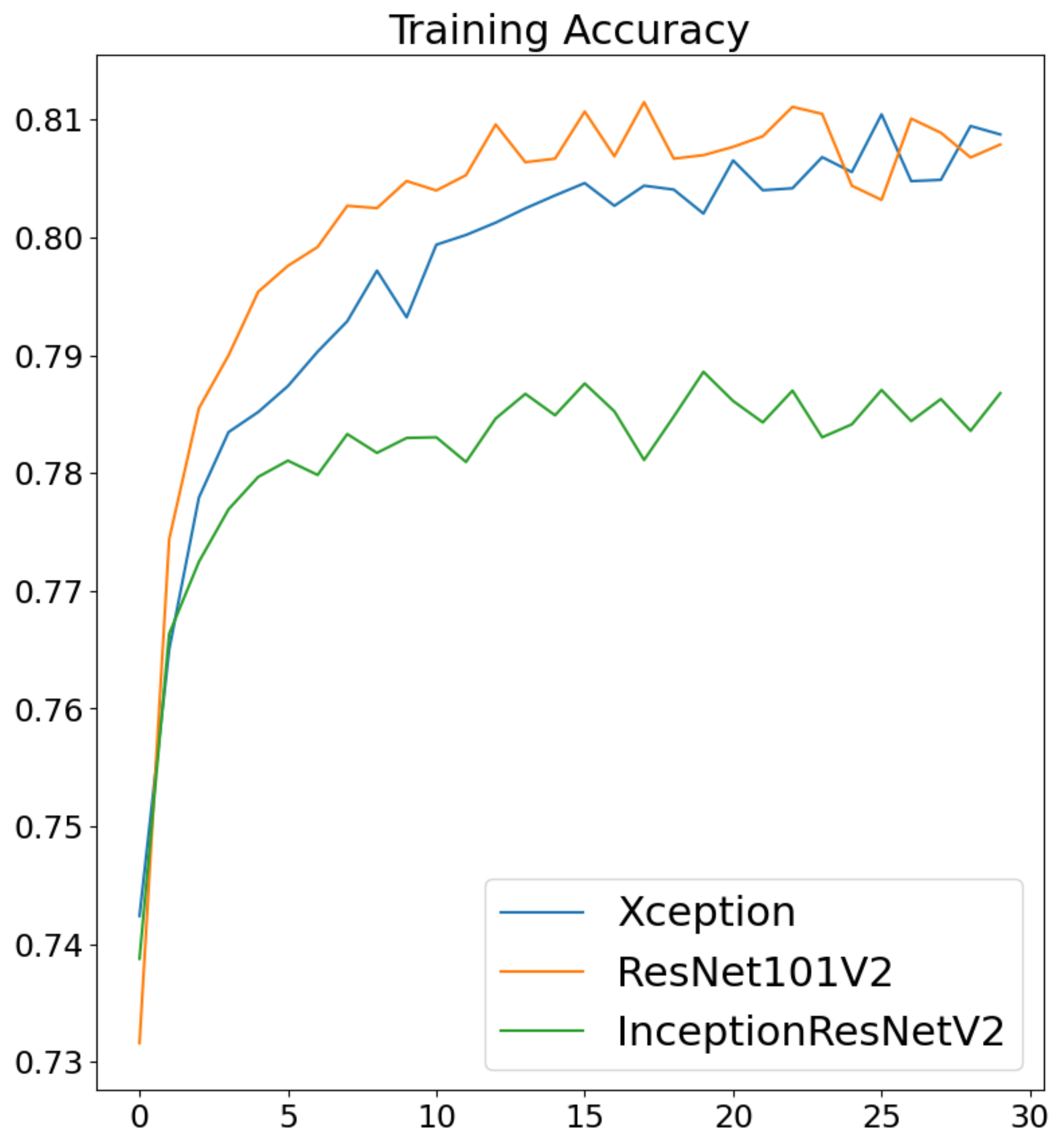}
         \subcaption{Training Accuracy}
      \label{fig_trainAcc}
     \end{subfigure}
     \hspace{1cm}
     \begin{subfigure}[b]{0.4\textwidth}
         \centering
         \includegraphics[width=\textwidth]{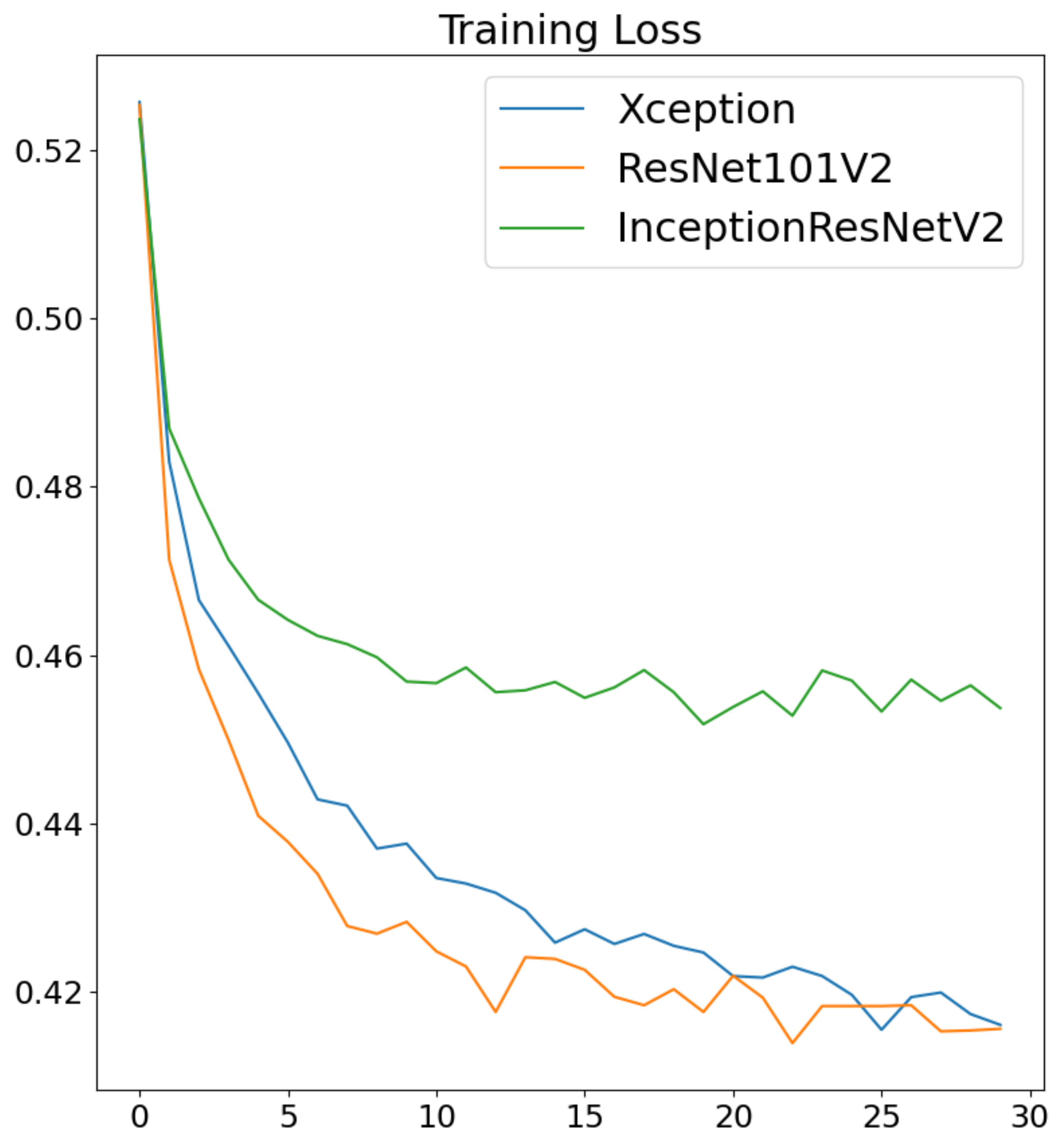}
         \subcaption{Training Loss}
      \label{fig_trainLoss}
     \end{subfigure}

        \caption{Results of feature extraction accuracy and loss for the transfer learning models}
        \label{fig:train_results}
\end{figure}

The summary of the feature extraction accuracy and loss during the training phase is shown in Figure \ref{fig:train_results}. 

% \begin{figure}[h!]
% \begin{minipage}[c]{.45\linewidth}
%   \centering
%     \includegraphics[scale=0.25]{figs/Accuracy.png}
%       \subcaption{Training Accuracy}
%       \label{fig_trainAcc}
% \end{minipage}\hfill
% \begin{minipage}[c]{.45\linewidth}
%   \centering
%     \includegraphics[scale=0.25]{figs/loss.png}
%       \subcaption{Training Loss}
%       \label{fig_trainLoss}
% \end{minipage}
% \caption{Results of feature extraction accuracy and loss for the transfer learning models}\label{fig:train_results}
% \end{figure}

As can be seen from Figure \ref{fig:train_results}, the accuracy of the Xception (network depth level $\sim$ 100) and ResNet101V2 (network depth level $\sim$ 200)  models increased rapidly and converged faster than the InceptionResNetV2 (network depth level $\sim$ 400) model during the training phase, while the training loss of the Xception and ResNet101V2 models drop significantly faster than the training loss of the InceptionResNetV2. This phenomenon is interesting because although the InceptionResNetV2 model had the highest depth level of the network architecture, it performed poorer than the other two models.

After fine-tuning for better performance, we evaluate the results of the models on the test dataset and present the results in Table \ref{tab:test_table}. We can observe that the ResNet101V2 model had the highest accuracy on the testing dataset at 95.72\%. Also, it had the highest Recall, F1 Score, and AUC score among the three models. The model with the best Precision is InceptionResNetV2. Also, InceptionResNetV2 performs slightly better than Xception on the test data. \\

\begin{table}[H]
    \centering
    \begin{tabular}{l|c|lllll}
\hline\hline
\textbf{Model} & \textbf{Network Depth Level}           & \textbf{Accuracy} & \textbf{Precision} & \textbf{Recall} & \textbf{F1 Score} & \textbf{AUC}    \\
\hline
Xception & $\sim 100$       & 0.9502   & 0.9436    & 0.9577 & 0.9506   & 0.9746 \\
ResNet101V2 & $\sim 200$     & \textbf{0.9572}   & 0.9505    & \textbf{0.9647} & \textbf{0.9575}   & \textbf{0.9758} \\
InceptionResNetV2 & $\sim 400$ & 0.9542   & \textbf{0.9541}    & 0.9541 & 0.9541   & 0.9712\\
\hline
\end{tabular}

    \caption{Performance of transfer learning models on testing data}
    \label{tab:test_table}
\end{table}

% We evaluate the performance of our approach against a single-shot   multibox detector (SSD). For the evaluation of our approach, the source image is split into a 10x10 grid producing 100 new images. Each new image is manually inspected for the presence of defects and assigned to their corresponding class, that is, 0 for non-defective and 1 for defective. We compare the performance of both approaches and illustrate the results in Table 2. \\

% \begin{table}[h!]
%     \centering
%     \input{contents/tables/table2.tex}
%     \caption{Comparison of model performance against SSD}
%     \label{tab:ssd_compare}
% \end{table}

\subsubsection{Evaluation on the Target Dataset}
For the evaluation of our approach on the unseen target dataset, an image is split into a 10x10 grid producing 100 new images. Each new image is manually inspected for the presence of defects and assigned to their corresponding class, that is, 0 for non-defective and 1 for defective. We compare the performance of the pre-trained models and illustrate the results in Table \ref{tab:target_compare}
We observe that the Xception model performs better than the two other models with an accuracy of 91.00\% and also a better Precision, F1 Score, and AUC. Xception is only surpassed by Resnet101V2 in Recall. 

\begin{table}[h!]
    \centering
    \begin{tabular}{l|c|lllll}
\hline\hline
\textbf{Model} & \textbf{Network Depth Level}          & \textbf{Accuracy} & \textbf{Precision} & \textbf{Recall} & \textbf{F1 Score} & \textbf{AUC} \\
\hline
Xception &$\sim 100$       & \textbf{0.9100}                       & \textbf{0.5385}                        & 0.7000                     & \textbf{0.6087}                       & \textbf{0.9128}                  \\
Resnet101V2 &$\sim 200$    & 0.7800                       & 0.2857                        & \textbf{0.8000}                     & 0.4210                       & 0.8883                  \\
InceptionResNetV2 &$\sim 400$ & 0.8700                       & 0.3846                        & 0.5000                     & 0.4347                       & 0.7833      \\
\hline
\end{tabular}

    \caption{Comparison of model performance on new unseen target object}
    \label{tab:target_compare}
\end{table}

\subsection{Defect Detection}
\begin{figure}[!h]
\begin{minipage}[c]{.33\linewidth}
  \centering
    \includegraphics[scale=0.095]{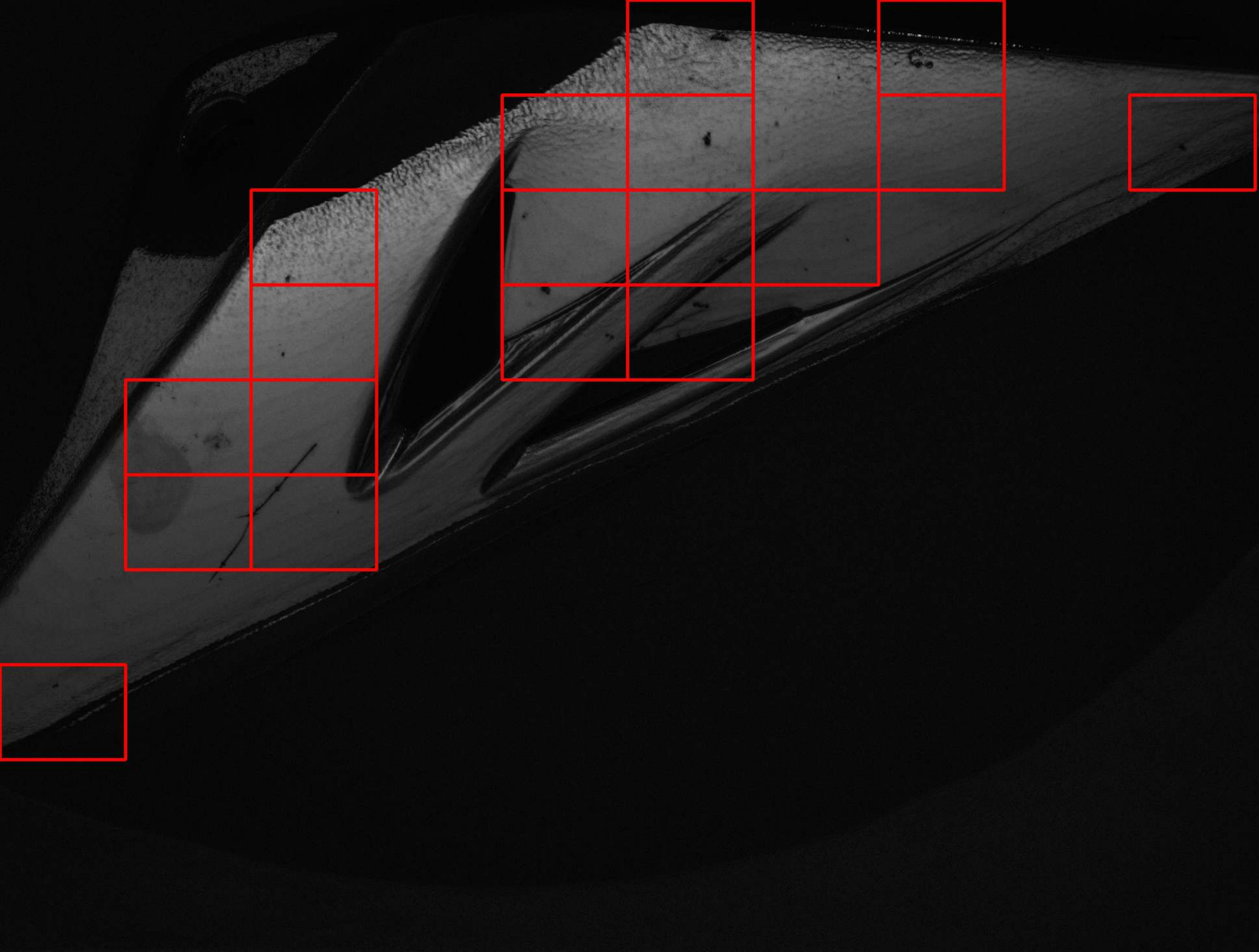}
      \subcaption{Xception}
      \label{fig_6a}
    \includegraphics[scale=0.095]{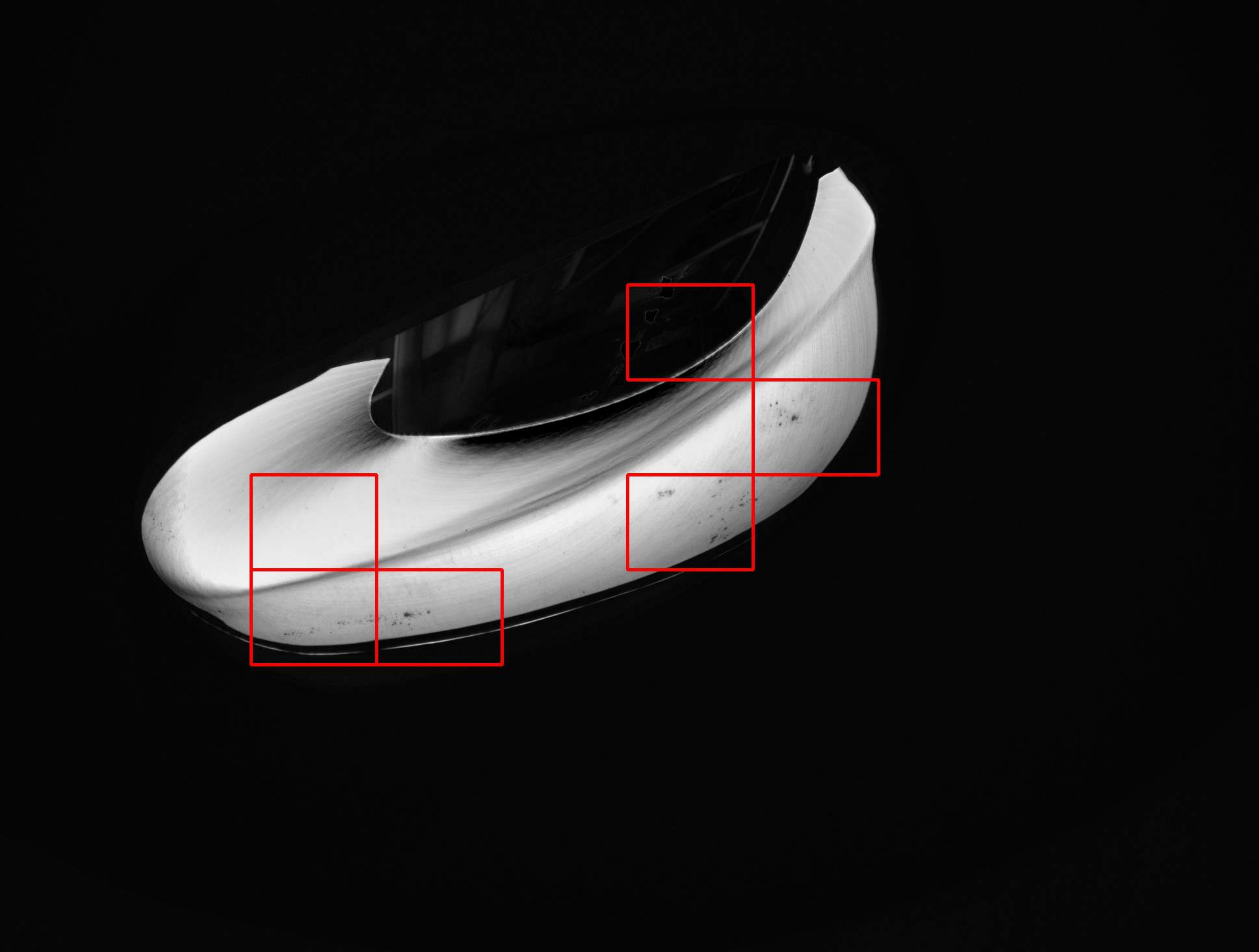}
      \subcaption{Xception}
      \label{fig_6b}
\end{minipage}\hfill
\begin{minipage}[c]{.33\linewidth}
  \centering
    \includegraphics[scale=0.095]{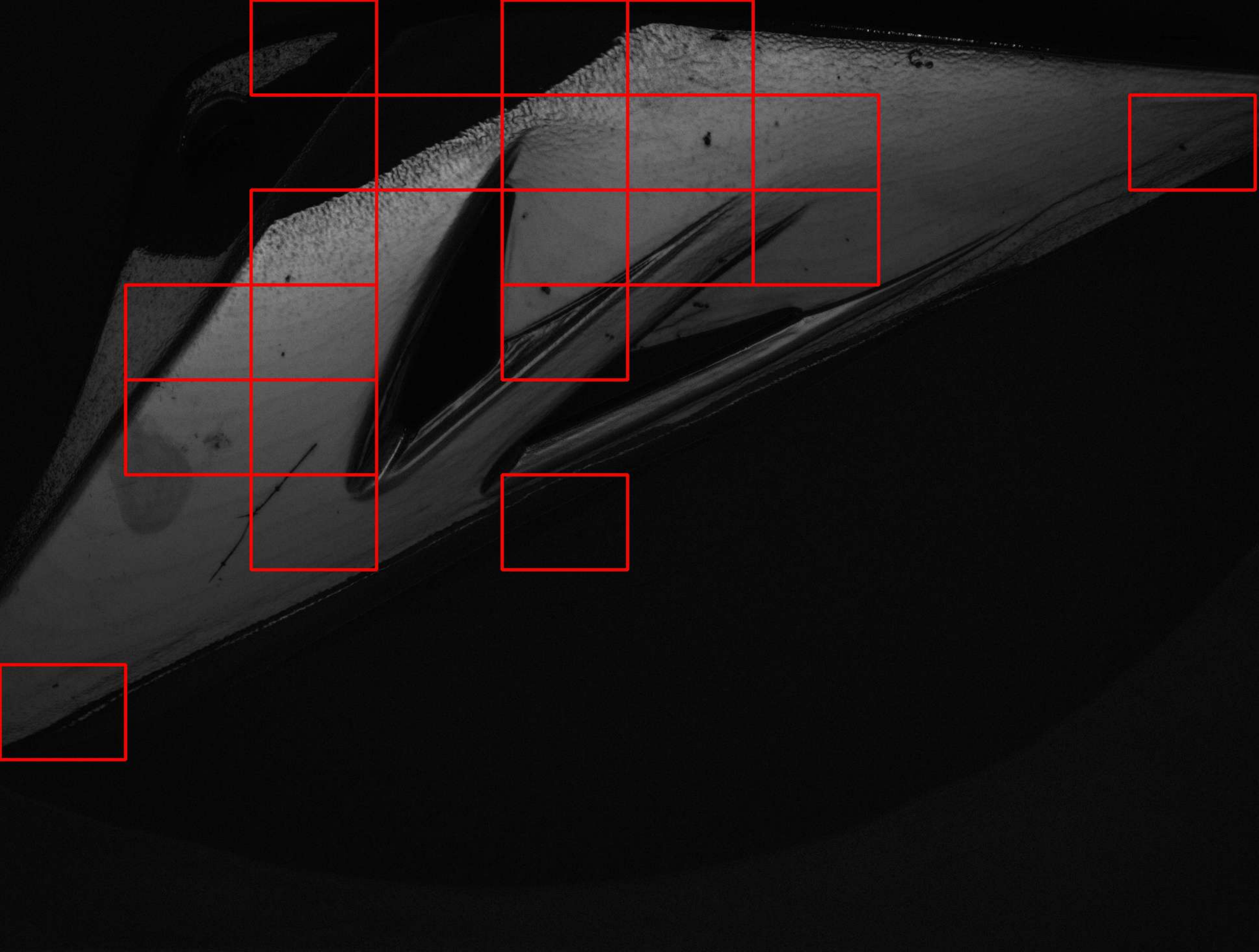}
      \subcaption{ResNet101V2}
      \label{fig_6c}  
    \includegraphics[scale=0.095]{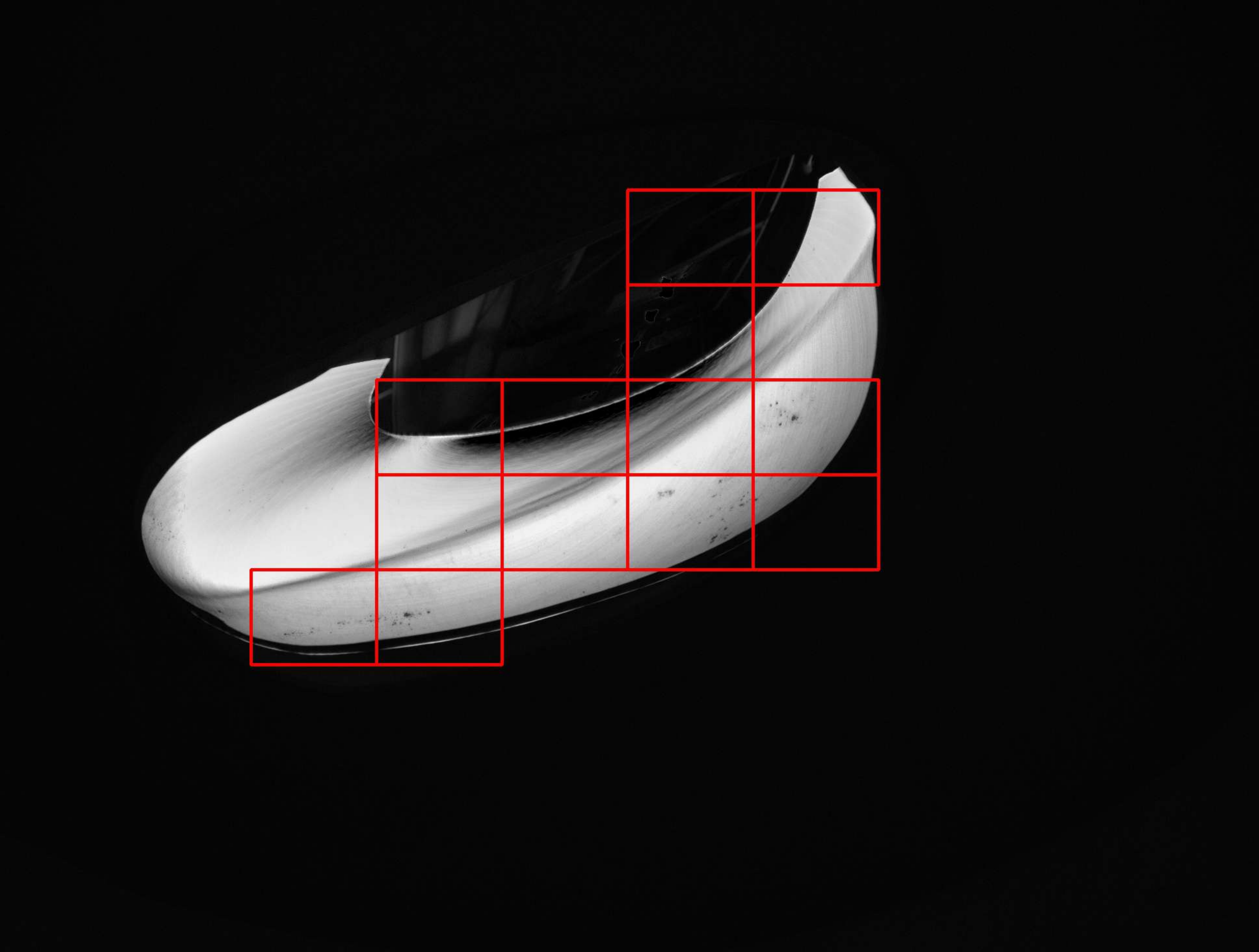}
      \subcaption{ResNet101V2}
      \label{fig_6d}  
\end{minipage}\hspace{0.001cm}%\hfill
\begin{minipage}[c]{.338\linewidth}
  \centering
    \includegraphics[scale=0.095]{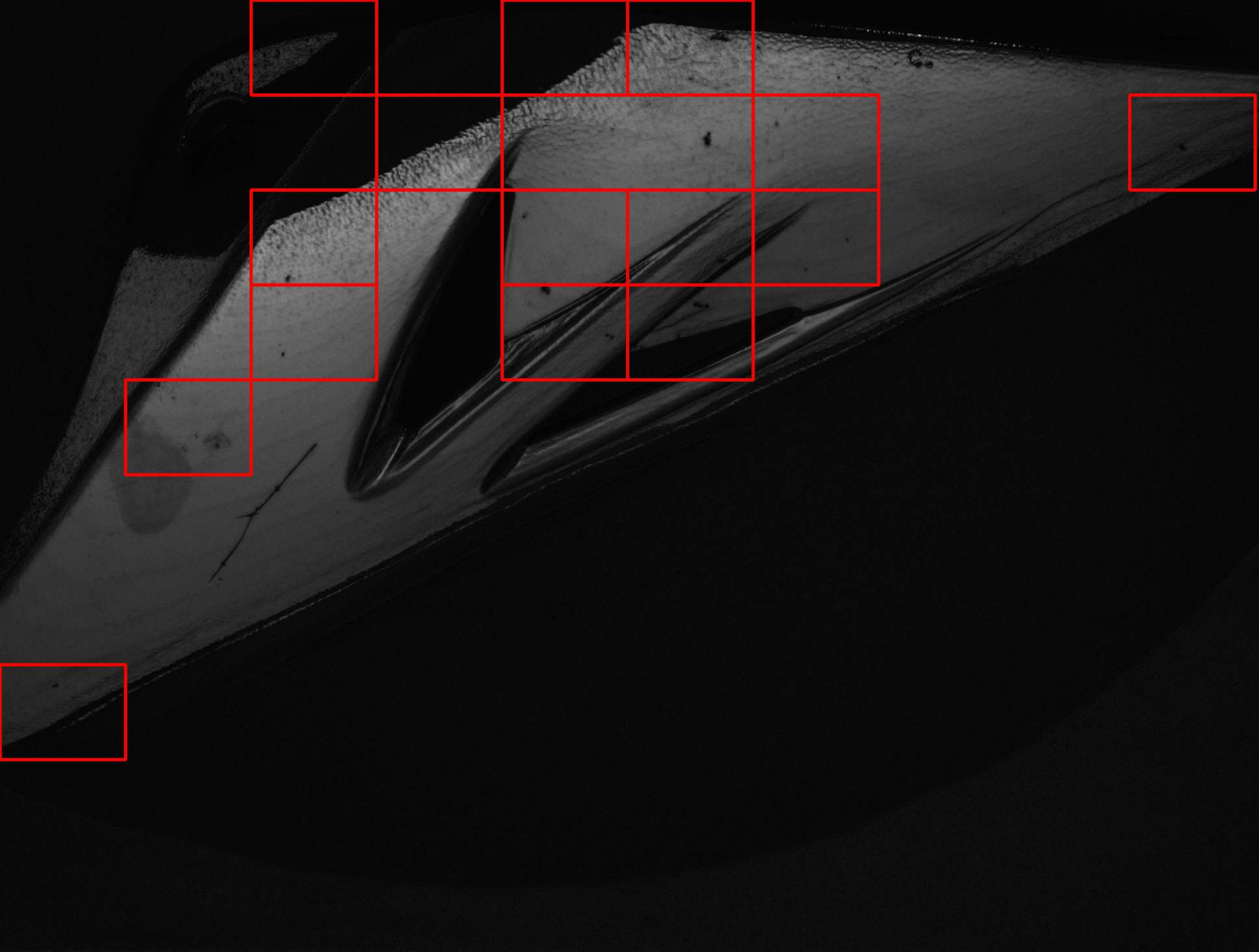}
      \subcaption{InceptionResNetV2}
      \label{fig_6e}
    \includegraphics[scale=0.095]{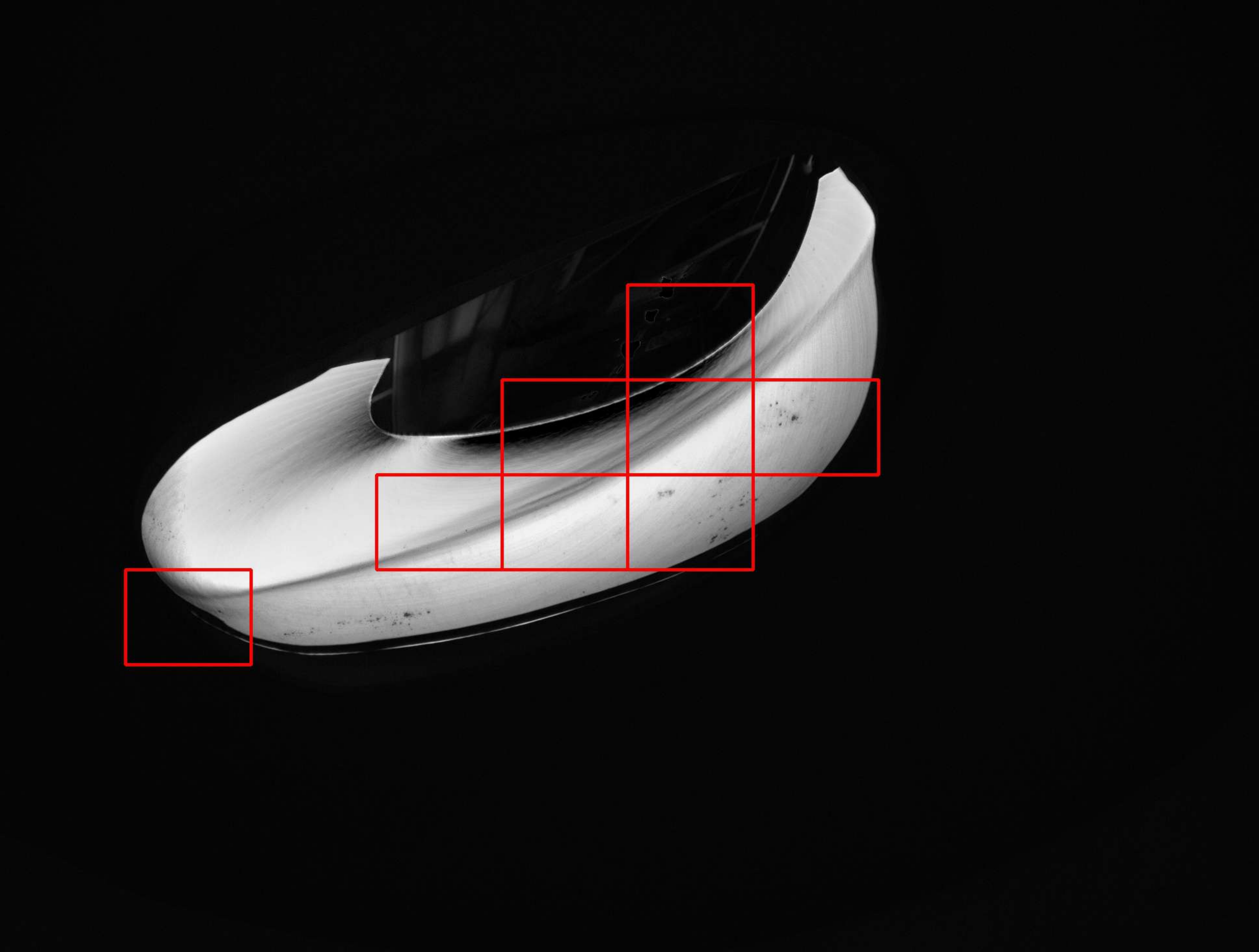}
      \subcaption{InceptionResNetV2}
      \label{fig_6f}
\end{minipage}

\caption{Result of defect detection using three different models on two new unseen target objects}\label{fig:detection_results}
\end{figure}

We discussed the defect detection phase in the previous Section \ref{proposed_defect_detection} and Section \ref{implemented_defect_detection}. We now evaluate the results of the defect detection on two images in our target dataset. We compare the results of the pseudo bounding boxes predicted by the transfer learning models on two new unseen target objects and present the results in Figure \ref{fig:detection_results}. 

The consistent performance of Xception on the unseen target objects is visible in the pseudo-object detection, where it correctly identifies more defective tiles (by visual inspection) than ResNet101V2 and InceptionResNetV2. Although InceptionResNetV2 has a better Accuracy, Precision, and F1 Score than ResNet101V2 (as seen in Table \ref{tab:target_compare}), in practice, ResNet101V2 performs better than InceptionResNetV2. The performance of InceptionResNetV2 is significantly poor on the defect detection.

%\subsection{Discussion}
 
We can conclude that the ability of the model to better extract features is crucial for its inference on the target images, irrespective of its performance after fine tuning. The results also justify the success of our data enhancement approach as both Xception and ResNet101V2 are able to extract relevant features about the source images from the enhanced dataset. Knowledge about these features is then transferred to the target dataset and used for inference on the new images.

%We can conclude that the ability of the model to better extract features is crucial for its inference on the target images, irrespective of its performance after fine tuning. The results also justify the success of our data enhancement approach as both Xception and Resnet101v2 are able to extract relevant features about the source images from the enhanced dataset. Knowledge about these features is then transferred to the target dataset and used for inference on the new images. 

From these results, we can also infer that the choice of the base model does not depend on its number of layers. 

\section{Conclusions}
Machine learning techniques have an important role in the smart manufacturing industry and have been well-embraced as a solution to many of the existing challenges in the domain. The identification of product imperfections is a crucial process in the manufacturing industry. Smart manufacturing makes use of machine learning techniques to facilitate efficient and rapid identification of these defects. However, the process is limited by the unavailability of large amounts of data required for model training.  

This paper proposed an architectural design that leverages Transfer Learning models for product defect detection applications with the goal of automating the quality assurance process in modern manufacturing. We also presented a method to enhance a small dataset to produce a sufficiently large dataset for training the models. We built an enhanced dataset of 22,700 labeled images from a source dataset containing 227 annotated images. We then used Transfer Learning models to build a robust classifier from the enhanced dataset that could accurately predict the presence of defects on industrial objects. We found that ResNet101V2 outperforms Xception and InceptionResNetV2 on our source data, with an accuracy of 95.72\%. However, Xception produced better results (accuracy of 91.00\%) than the other two models when tested on unseen target objects. Our results also show that the choice of a pre-trained model is not dependent on the depth of the network.

The success of our approach is justified by the significant performance on both the source dataset and a target dataset containing unseen and unlabelled images. This proves that our proposed method can be implemented on defect detection applications with limited source data available, and the knowledge learned can be extended to new unseen data. Our proposed approach is currently limited to the datasets discussed in this paper. However, future work can be extended by applying the method to different defect datasets and evaluating the results.  

\section*{Acknowledgements}
This work is supported by the NBIF Pre-AI Voucher Fund (AIP2023-007).

%Begin appendix section(s)
%\appendix

% % Add appendices here:
% \section{Example of math equation }
% %\label{appendix-customize-this-label}
% Binomial theorem: \cite{abramowitz1948handbook}
% \begin{equation}
% (x+y)^n=\sum_{\substack{k=0}}^{n}\dbinom{n}{k}x^{n-k}y^k
% \end{equation}

% All references should be stored in the file "references.bib"
% Please do not modify anything below this line.
%\thispagestyle{plain}
%\printbibliography

\printbibliography[heading=subbibintoc]

\end{document}